\pdfoutput = 1
\documentclass[10pt,twocolumn,letterpaper]{article}
\usepackage{array}
\usepackage{colortbl}
\usepackage{amsmath}
\usepackage{cvpr}
\usepackage{times}
\usepackage{epsfig}
\usepackage{overpic}
\usepackage{graphicx}
\usepackage{amssymb}
\usepackage{amsmath}
\usepackage{subfigure}
\usepackage{amssymb,bm}
\usepackage{tabulary}
\usepackage{microtype}
\usepackage{booktabs}
\usepackage[table]{xcolor}
\usepackage{multirow}
\usepackage{soul}
\usepackage[english]{babel}
\usepackage[numbers,sort]{natbib} 

\usepackage[pagebackref=true,breaklinks=true,letterpaper=true,colorlinks,bookmarks=false]{hyperref}

\cvprfinalcopy 

\def\cvprPaperID{****} 

\begin{document}

\title{Parsing R-CNN for Instance-Level Human Analysis}

\author{Lu Yang$^1$, Qing Song$^1$, Zhihui Wang$^1$ and Ming Jiang$^2$\\
$^1$Beijing University of Posts and Telecommunications, $^2$WiWide Inc.\\
{\tt\small $^1$\{soeaver, priv, wangzh\}@bupt.edu.cn, $^2$ming@wiwide.com}
}

\maketitle

\begin{abstract}
   Instance-level human analysis is common in real-life scenarios and has multiple manifestations, such as human part segmentation, dense pose estimation, human-object interactions, etc. Models need to distinguish different human instances in the image panel and learn rich features to represent the details of each instance. In this paper, we present an end-to-end pipeline for solving the instance-level human analysis, named Parsing R-CNN. It processes a set of human instances simultaneously through comprehensive considering the characteristics of region-based approach and the appearance of a human, thus allowing representing the details of instances. 
   
   Parsing R-CNN is very flexible and efficient, which is applicable to many issues in human instance analysis. Our approach outperforms all state-of-the-art methods on CIHP (Crowd Instance-level Human Parsing), MHP v2.0 (Multi-Human Parsing) and DensePose-COCO datasets. Based on the proposed Parsing R-CNN, we reach the 1st place in the COCO 2018 Challenge DensePose Estimation task. Code and models are public available\footnote{\fontsize{7pt}{1em}\url{https://github.com/soeaver/Parsing-R-CNN}}.
\end{abstract}


\section{Introduction}

Human part segmentation~\cite{Liang_tpami2018_lip, Zhao_mm2018_mhpv2, Li_mm2018_mhpm, Gong_eccv2018_pgn}, dense pose estimation~\cite{Guler_cvpr2018_densepose} and human-object interactions~\cite{Hu_cvpr2018_relation, Gkioxari_cvpr2018_interacnet, Gao_bmvc2018_ican} are the most fundamental and critical tasks in analyzing human in the wild. These tasks require human details at the instance level, which involve several perceptual tasks including detection, segmentation, estimation, i.e. There is a commonality between them, which can be regarded as an instance-level human analysis task. 

\begin{figure}[t]
\begin{center}
\includegraphics[height=65mm]{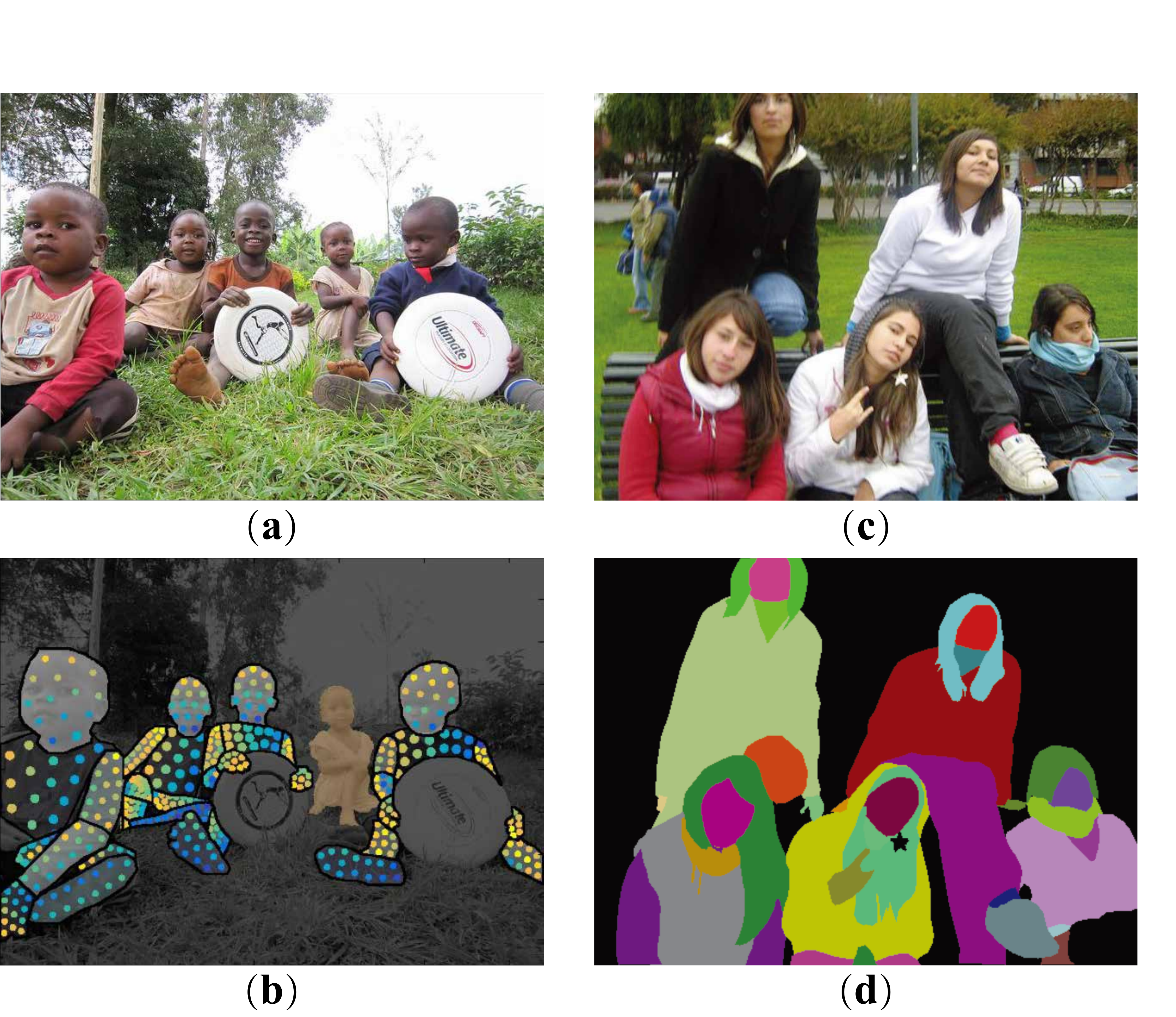}
\end{center}
\caption{Example tasks of instance-level human analysis. (\textbf{a}) and (\textbf{b}) are samples for dense pose estimation. (\textbf{c}) and (\textbf{d}) are samples for human part segmentation.}
\label{fig:instance_level_human_analysis}
\end{figure}

Due to the successful development of convolutional neural networks~\cite{Alex_nip2012_alexnet, Girshick_cvpr2014_rcnn, Long_cvpr2015_fcn, Szegedy_cvpr2016_inception-v3, He_iccv2017_maskrcnn, Wu_eccv2018_gn}, great progress has been made in instance-level human analysis, especially in human part segmentation and dense pose estimation. Several related works~\cite{Liang_tpami2018_lip, Zhao_mm2018_mhpv2} follow the two stages pipeline, Mask R-CNN~\cite{He_iccv2017_maskrcnn}, which detects human in the image panel and predicts a class-aware mask in parallel with several convolutional layers. This method has achieved great success and wide application in instance segmentation~\cite{Dai_cvpr2016_mnc, Dai_cvpr2017_fcis, He_iccv2017_maskrcnn, Liu_cvpr2018_panet}. However, there are still several deficiencies in extending to the instance-level human analysis. One of the most important problems is that the design of the mask branch is used to predict a class-agnostic instance mask~\cite{He_iccv2017_maskrcnn}, but the instance-level human analysis requires more detailed features, which can not be well solved by existing methods. Besides, human analysis needs to correlate geometric and semantic relations between human parts / dense points, which is also missing. Therefore, in order to solve these problems, we propose Parsing R-CNN, which provides a concise and effective scheme for the instance-level human analysis tasks. This scheme can be successfully applied to the human part segmentation and dense pose estimation (Figure~\ref{fig:instance_level_human_analysis}). 

\begin{figure*}
\begin{center}
\includegraphics[width=0.89\linewidth]{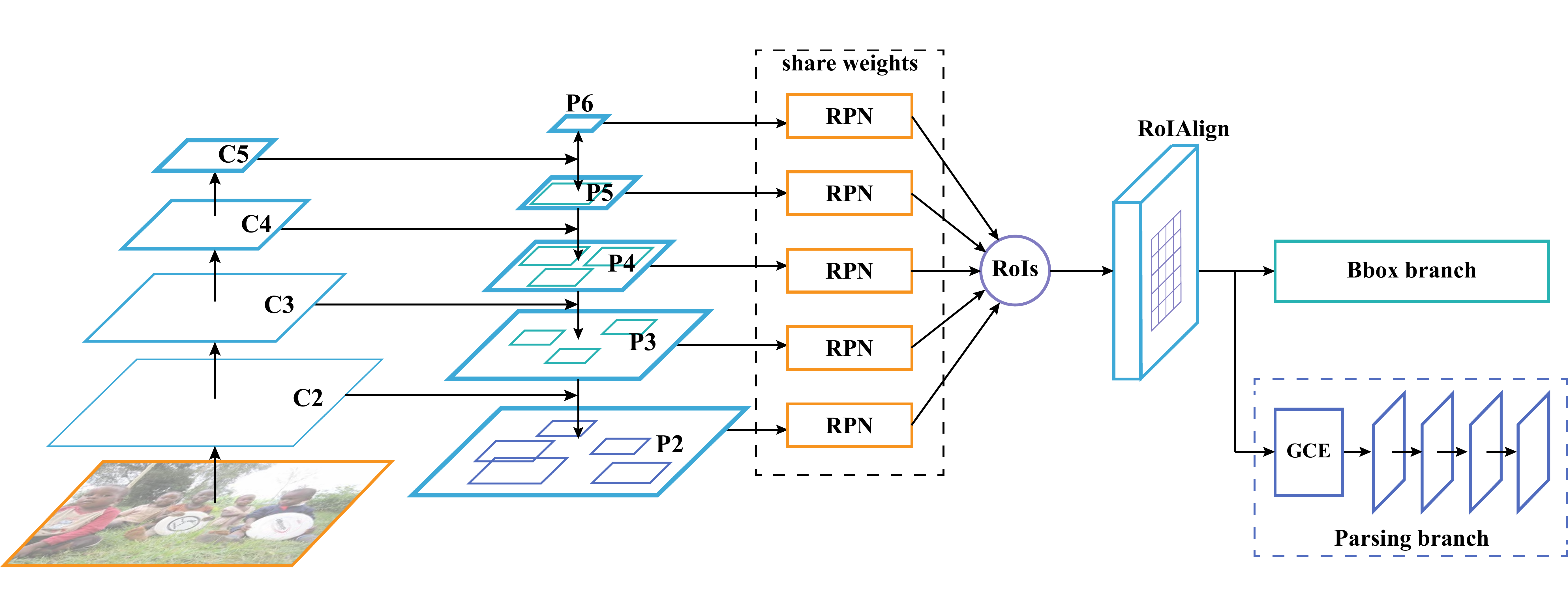}
\end{center}
\vspace{-1mm}
\caption{Parsing R-CNN pipeline. We adopt FPN backbone and RoIAlign operation, parsing branch is used for instance-level human analysis.}
\label{fig:parsing_rcnn_pipline}
\end{figure*}

Our research explores the problem of instance-level human analysis from four aspects. First, to enhance feature semantic information and maintain feature resolution, proposals separation sampling is adopted. Human instances often occupy a relatively large proportion in images~\cite{Lin_eccv2014_coco}. Therefore, RoIPool~\cite{Girshick_iccv2015_fast-rcnn} operations are often performed on the coarser-resolution feature maps~\cite{Lin_cvpr2017_fpn}. But this will lose a lot of details of the instance. In this work, we adopt the proposals separation sampling strategy, which using pyramid features at RPN~\cite{Ren_nips2015_faster-rcnn} phase, but the RoIPool only performed on the finest level.

Second, to obtain more detailed information to distinguish different human parts or dense points in the instance, we enlarge the RoI resolution of the parsing branch. Human analysis tasks generally distinguish between dozens or even dozens of categories. It is necessary and effective to enlarge the resolution of the feature map.

Third, we propose a geometric and context encoding module to enlarge receptive field and capture the relationship between different parts of the human body. It is a lightweight component consisting of two parts. The first part is used to obtain multi-level receptive field and context information, and the second part is used to learn geometric correlation. With this module, class-aware masks with better quality are produced.

Finally,  to explore the functions of each group operation in parsing, we decouple the branch into three parts: semantic space transformation, geometric and context encoding, semantic feature representation. Meanwhile, we propose an appropriate branch composition scheme with high accuracy and small computational overhead.

With the proposed Parsing R-CNN, we achieve state-of-the-art performance on several datasets~\cite{Zhao_mm2018_mhpv2, Gong_eccv2018_pgn}. For human part segmentation, Parsing R-CNN outperforms all known top-down or bottom-up methods both on CIHP~\cite{Gong_eccv2018_pgn} (Crowd Instance-level Human Parsing) and MHP v2.0~\cite{Zhao_mm2018_mhpv2} (Multi-Human Parsing) datasets. For dense pose estimation, Parsing R-CNN achieves 64.1\% mAP on COCO DensePose~\cite{Guler_cvpr2018_densepose} test dataset, winning the 1st place in COCO 2018 Challenge DensePose task by a very large margin.


Parsing R-CNN is general and not limited to human part segmentation and dense pose estimation. We do not see any reason preventing it from finding broader applications in other human analysis tasks, such as human-object interactions, etc.

\section{Related Work}

\noindent\textbf{Region-based Approach.} The region-based approach~\cite{Girshick_cvpr2015_dpm, Girshick_cvpr2014_rcnn, He_eccv2014_sppnet, Girshick_iccv2015_fast-rcnn, Ren_nips2015_faster-rcnn, Lin_cvpr2017_fpn, He_iccv2017_maskrcnn} is a very important series in object detection, which has high accuracy and good expansibility. Generally speaking, the region-based approach generates a series of candidate object regions~\cite{Uijlings_ijcv2013_selective, Zitnick_eccv2014_edgebox, Ren_nips2015_faster-rcnn}, then performs object classification and bounding-box regression in parallel within each candidate region. RoIPool and Region Proposal Network (RPN) are proposed by Fast R-CNN~\cite{Girshick_iccv2015_fast-rcnn} and Faster R-CNN~\cite{Ren_nips2015_faster-rcnn} respectively, which enable the region-based approach end-to-end learning and greatly improve speed and accuracy. Mask R-CNN~\cite{He_iccv2017_maskrcnn} is an important milestone that successfully extending the region-based approach to instance segmentation and pose estimation, which has become an advanced pipeline in visual recognition. Mask R-CNN is flexible and robust to many follow-up improvements, and can be extended to more visual tasks~\cite{Cao_cvpr2017_openpose, Guler_cvpr2018_densepose, Gkioxari_cvpr2018_interacnet, Rohit_cvpr2018_dandt, Hu_cvpr2018_segment}.

\vspace{6pt}
\noindent\textbf{Human Part Segmentation.} Human part segmentation is a core task of human analysis, which has been extensively studied in recent years. Recently, Zhao \etal~\cite{Zhao_mm2018_mhpv2} put forward the MHP v2.0 (Multi-Human Parsing) dataset, which contains 25,403 elaborately annotated images with 58 fine-grained semantic category labels. Gong \etal~\cite{Gong_eccv2018_pgn} present another large-scale dataset called Crowd Instance-level Human Parsing (CIHP) dataset, which has 38,280 diverse human images. Each image in CIHP is labeled with pixel-wise annotations on 20 categories and instance-level identification. These datasets have greatly promoted the research of human part segmentation, and considerable progress has been made. 

On the other hand, Zhao \etal~\cite{Zhao_mm2018_mhpv2} propose the Nested Adversarial Network (NAN) for human part segmentation, which consists of three GAN-like sub-nets, respectively performing semantic saliency prediction, instance-agnostic parsing and instance-aware clustering. Gong \etal~\cite{Gong_eccv2018_pgn} design a detection-free Part Grouping Network (PGN) for instance-level human part segmentation. Although these works have achieved good performance, the segmentation result has great room for improvement and lack of an efficient end-to-end pipeline to unify the solution of instance-level human analysis.

\vspace{6pt}
\noindent\textbf{Dense Pose Estimation.} Guler \etal~\cite{Guler_cvpr2018_densepose} propose an innovative dataset for instance-level human analysis, DensePose-COCO, a large-scale ground-truth dataset with image-to-surface correspondences manually annotated on 50k COCO images. Dense pose estimation can be understood as providing a refined version of human part segmentation and pose estimation, where one predicts continuous part labels of each human body. They also present the DensePose-RCNN, which combines the Dense Regression approach with the Mask-RCNN~\cite{He_iccv2017_maskrcnn} architecture. Cross-cascading architecture is applied to the system that further improves accuracy. DensePose-RCNN gives a concise pipeline for dense pose estimation with good accuracy. However, many problems in the task are not discussed, such as the scale of human instance, the feature resolution and so on.

We consider that we can not treat human part segmentation and dense pose estimation in isolation. They are both specific tasks of instance-level human analysis and have a lot of commonalities. Therefore, based on the successful region-based approach, we propose Parsing R-CNN, a unified solution for instance-level human analysis.

\section{Parsing R-CNN}
\label{sec:prcnn}

Our goal is to leverage a unified pipeline for instance-level human analysis, which can achieve good performance in both human part segmentation, dense pose estimation and has the high scalability to other similar tasks~\cite{Gkioxari_cvpr2018_interacnet, Rohit_cvpr2018_dandt}. Like Mask R-CNN, the proposed Parsing R-CNN is conceptually simple, an additional parsing branch is used to generate the output of instance-level human analysis, as shown in Figure~\ref{fig:parsing_rcnn_pipline}. In this section, we will introduce the motivation and content of Parsing R-CNN in detail.

\subsection{Proposals Separation Sampling}

\begin{figure}[t]
\begin{center}
\includegraphics[height=50mm]{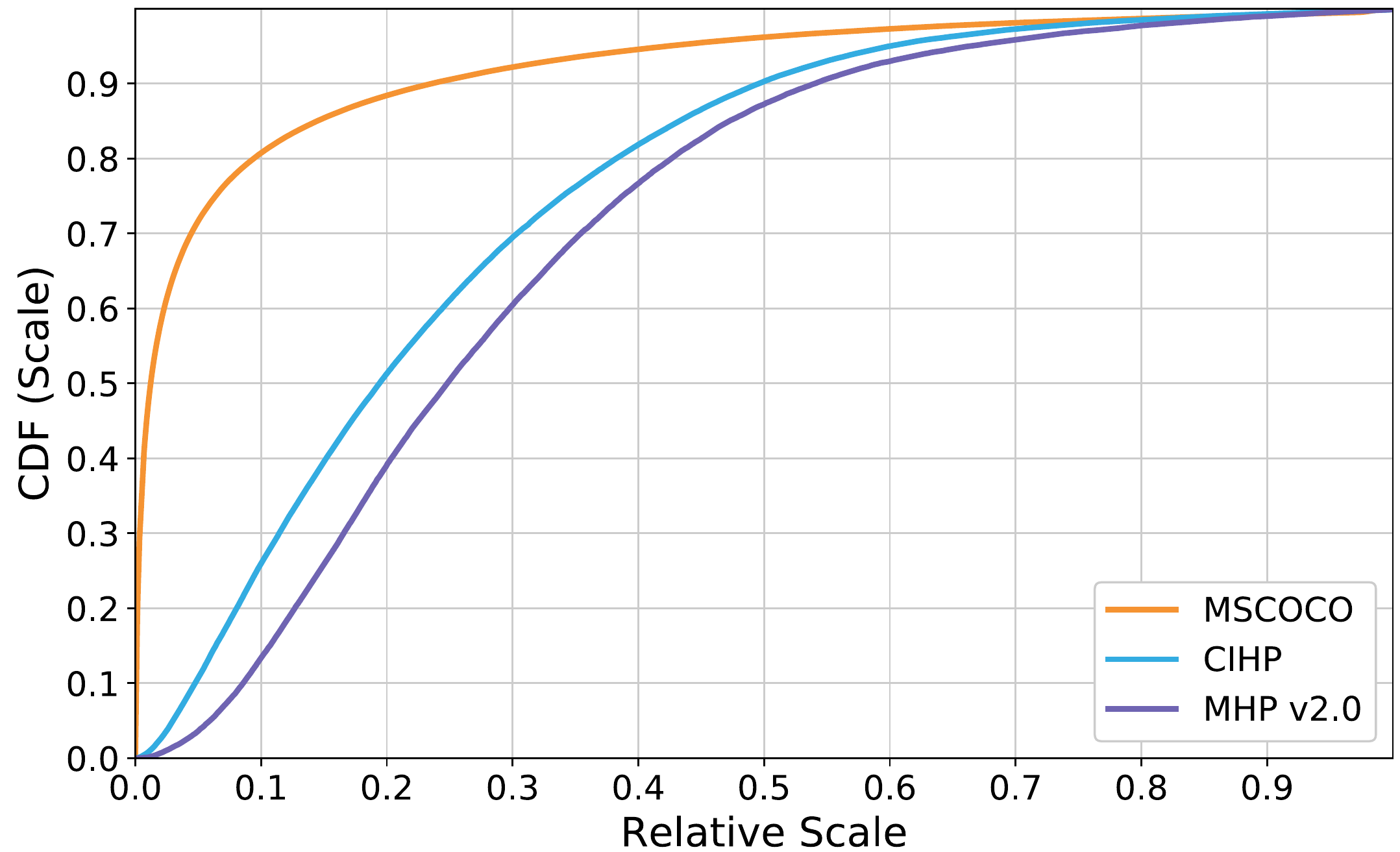}
\end{center}
\caption{Scale of instances relative to the image (\textbf{Relative Scale})  $vs$ fraction of instances in the dataset (\textbf{CDF}).}
\label{fig:scale_ratio}
\end{figure}

In FPN~\cite{Lin_cvpr2017_fpn} and Mask R-CNN~\cite{He_iccv2017_maskrcnn}, the assignment strategy is adopted to collect the RoIs (Regions of Interest) and assign them to the corresponding feature pyramid according to the scale of RoIs. Formally, large RoIs will be assigned to the coarser-resolution feature maps. This strategy is effective and efficient in object detection and instance segmentation. However, we find that this strategy is not the optimal solution in instance-level human analysis. Due to a small instance cannot be accurately annotated as part segmentation or dense pose, human instances often occupy a larger scale of the image. As shown in Figure~\ref{fig:scale_ratio}, less than 20\% of object instances in COCO dataset occupy more than 10\% scale of the image, but this ratio is about 74\% and 86\% in CIHP and MHP v2.0 datasets respectively. According to the assignment strategy proposed by FPN, the most human instances will be assigned to the coarser-resolution feature maps. Instance-level human analysis often requires precise identification of some details of the human body, such as glasses and watches, or pixel areas of the left and right hand. But the coarser-resolution feature maps cannot provide more instance details, which is very harmful to human analysis.

To address this, we propose the \emph{proposals separation sampling} (PSS) strategy that extracts features with details while preserves a multi-scale feature representation~\cite{Lin_cvpr2017_fpn, Liu_eccv2016_ssd, Li_arxiv2017_fssd}. Our proposed change is simple: the bbox branch still adopts the scale assign strategy on the feature pyramid (\emph{P2-P5}) according to FPN~\cite{Lin_cvpr2017_fpn}, but the RoIPool/RoIAlign operation of parsing branch is only performed on the finest scale feature map of \emph{P2}, as shown in Figure~\ref{fig:parsing_rcnn_pipline}. In this way, we argue that object detection benefits from the pyramid representation while preserving human body details by extracting feature from the finest-resolution feature maps at parsing branch. With PSS, we observe that there has been a significant improvement in human part segmentation and dense pose estimation.

\subsection{Enlarging RoI Resolution}

\begin{figure}[t]
\begin{center}
\includegraphics[height=78mm]{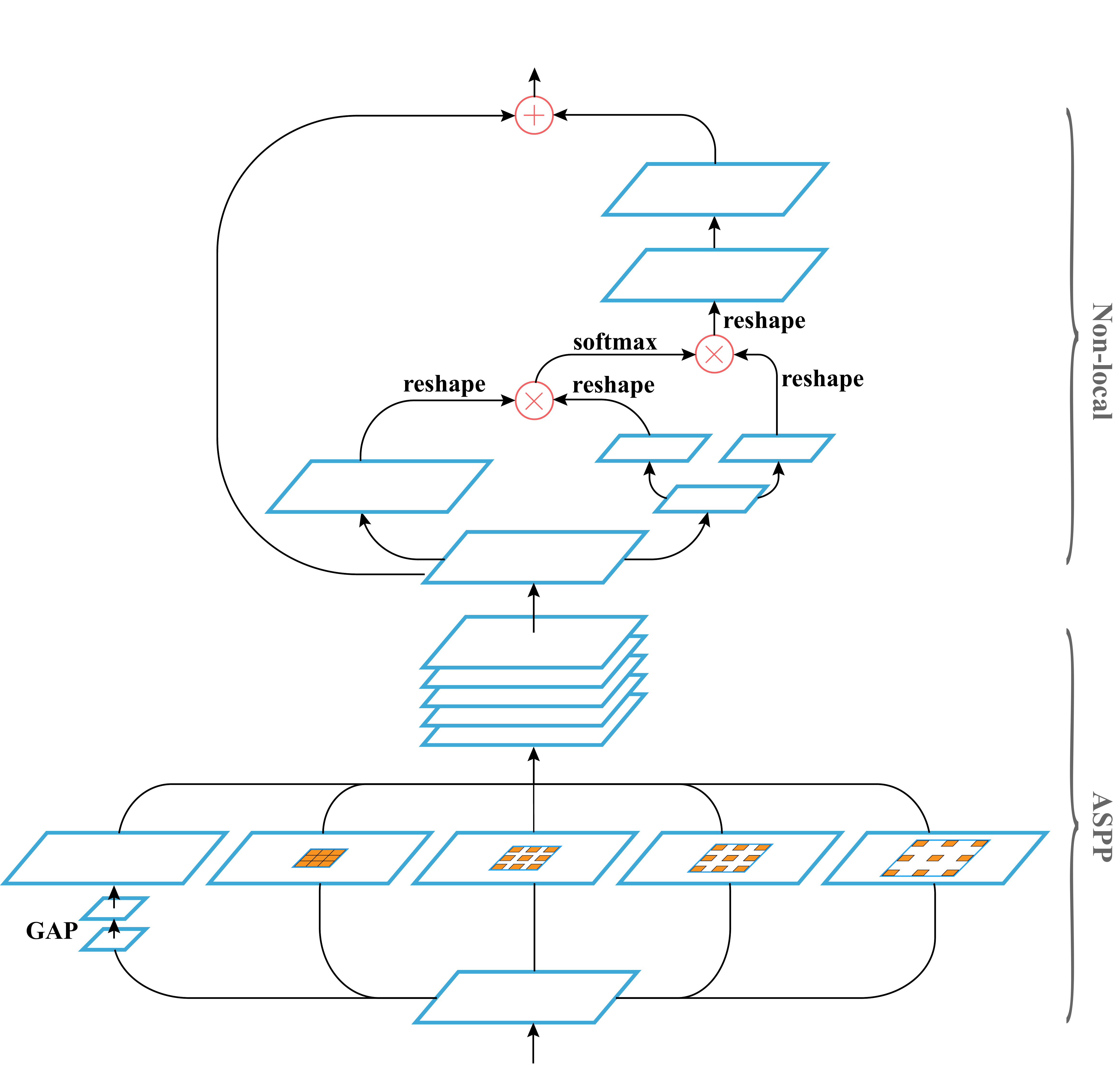}
\end{center}
\caption{The proposed  \emph{Geometric and Context Encoding} (GCE) module.}
\label{fig:gce_module}
\end{figure}

In some early region-based approaches, in order to make full use of the pre-train parameters, RoIPool operation converts an RoI into a small feature map with a fixed spatial extent of 7$\times$7~\cite{Girshick_iccv2015_fast-rcnn, Lin_cvpr2017_fpn, He_iccv2017_maskrcnn} (or 14$\times$14 followed by a convolutional layer with stride=2). This setup has been inherited in subsequent work and has proved its efficiency. Mask R-CNN uses 14$\times$14 scale RoIs in mask branch to generate segmentation masks, and the DensePose-RCNN~\cite{Guler_cvpr2018_densepose} uses the same settings in the uv branch. But the most human instances occupy a large proportion of the feature maps, and too small RoI will lose a lot of detail. For example, a 160$\times$64 size human body whose size on \emph{P2} is 40$\times$16, and scaling to 14$\times$14 will undoubtedly reduce the prediction accuracy. In the tasks of object detection and instance segmentation, it is not very necessary to accurately predict the details of the instance. But in the instance-level human analysis, this will cause severe accuracy degradation.

In this work, we present the most simple and intuitive method: \emph{enlarging RoI resolution} (ERR). We employ 32$\times$32 RoI in parsing branch, which increases the computational cost of the branch, but improves the accuracy significantly. To address the training time and memory overhead associated with ERR, we decoupled the batch size of instance-level human analysis tasks from the detection task to a fixed value (e.g. 32) and find that this greatly increases the training speed and does not lead to accuracy degradation.

\subsection{Geometric and Context Encoding}

\begin{figure}[t]
\begin{center}
\includegraphics[height=60mm]{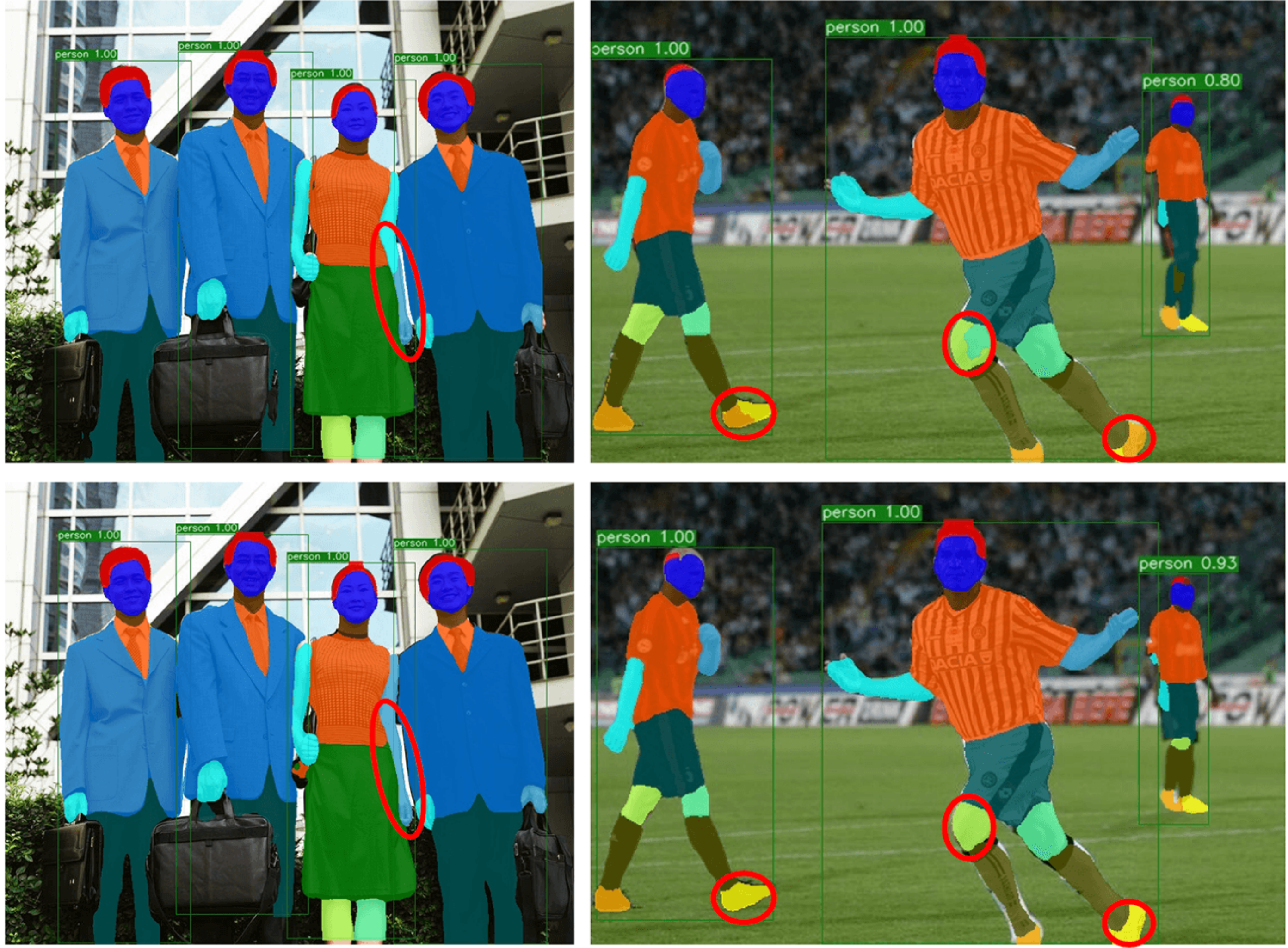}
\end{center}
\caption{Visualization results with / without GCE module. The 1st row shows visualization results without GCE, and the 2nd shows ones with GCE. The GCE module can refine segmentation results of human instances (red circles).}
\label{fig:with_gce}
\end{figure}

In previous works, the design of each branch is very succinct. A tiny FCN~\cite{Long_cvpr2015_fcn} is applied on the pooled feature grid for predicting pixel-wise masks of instances. However, using a tiny FCN in the parsing branch of instance-level human analysis will have three obvious drawbacks. First, the scale of different human parts varies greatly, which requires the feature maps capturing multi-scale information. Secondly, each human part is geometrically related, which requires a non-local representation~\cite{Buades_cvpr2005_nonlocal} . Third, 32$\times$32 RoI needs a large receptive field, and stacking four or eight 3$\times$3 convolutional layers are not enough. 

\emph{Atrous spatial pyramid pooling} (ASPP)~\cite{Chen_tpami2016_deeplab, Chen_arxiv2017_deeplabv3, Chen_eccv2018_deeplabv3plus} is an effective module in semantic segmentation, where parallel atrous convolutional layers with different rates capture multi-scale information. Recently, Wang \etal presents the non-local operation and demonstrates outstanding performance on several benchmarks. Non-local~\cite{Wang_cvpr2018_nonlocal} operation is able to capture long-range dependencies which is of central importance in deep neural networks. For instance-level human analysis, we combine the advantages of ASPP and non-local, propose the \emph{Geometric and Context Encoding} (GCE) module to replace FCN in parsing branch. As shown in Figure~\ref{fig:with_gce}, the proposed GCE module can encode the geometric and context information of each instance, effectively distinguish different parts of the human body. In the GCE module, the ASPP part consists of one 1$\times$1 convolution and three 3$\times$3 convolutions with rates = (6, 12, 18). The image-level features are generated by global average pooling, which is followed by a 1$\times$1 convolution, and then bilinearly upsample the feature to the original 32$\times$32 spatial dimension. The non-local part adopts embedded Gaussian version, and a batch normalization~\cite{Ioffe_icml2015_bn} layer is added to the last convolutional layer. All the convolutional layers in GCE module have 256 channels. See Figure~\ref{fig:gce_module}.

\subsection{Parsing Branch Decoupling}
In the design of neural network for visual task, we often divide the network into several parts according to the characteristics of the features learned by different convolutional layers. For example, high layers strongly respond to entire objects while other neurons are more likely to be activated by local texture and patterns. The region-based approach handles each RoI in parallel, so the branch of each task can be understood as an independent neural network. However, the existing works have not decoupled the branch into different parts and analyzed their roles.

In this work, we decouple the parsing branch for instance-level human analysis into three parts. We consider that each part plays a different role for the task. The first part is for semantic space transformation, which is used to transform features into corresponding tasks. The second part is GCE module for geometric and context encoding. The last part converts semantic features to specific tasks, and can also be used to enhance the network capacity. We call them \emph{before GCE}, \emph{GCE module} and \emph{after GCE} respectively. For instance-level human analysis, it is not simple to increase the computational complexity of each module to improve the accuracy. Therefore, it is necessary to decouple parsing branch and analyze the speed / accuracy trade-offs of each part.

\section{Experiments}
\label{sec:exper}

In this section, we compare the performance of Parsing R-CNN on three datasets, two human part segmentation datasets, and one dense pose estimation dataset.

\begin{table}[t]
\centering
\small
\begin{tabular}{c|c|cccc}
 & mAP$^\text{bbox}$  & mIoU  &  AP$^\text{p}_\text{50}$ & AP$^\text{p}_\text{vol}$  & PCP$_\text{50}$ \\
 \toprule[0.2em]
baseline & \textbf{67.7}  &47.2  & 41.4 & 45.4 & 44.3 \\
P2 only  & 66.4  & 47.7 & 42.6 & 45.8 & 45.1 \\
PSS       & 67.5  &  \textbf{48.2} &  \textbf{42.9}  &  \textbf{46.0}  &  \textbf{45.5}  \\

\end{tabular}
\vspace{.5em}
  \caption{Ablation study on \emph{proposals separation sampling} (PSS) strategy.}
  \label{tab:ablation_pss}
\vspace{-.8em}
\end{table}

\begin{table}[t]
\centering
\small
\scalebox{0.9}{
\begin{tabular}{c|c|cccc}
 & fps & mIoU  &  AP$^\text{p}_\text{50}$ & AP$^\text{p}_\text{vol}$  & PCP$_\text{50}$ \\
 \toprule[0.2em]
baseline (14$\times$14) & 10.4 & 48.2 & 42.9 & 46.0 & 45.5  \\
ERR (32$\times$32)       & 9.1   & 50.7 & 47.9 & 47.6 & 49.7 \\
ERR (32$\times$32), 100 RoIs  & \textbf{11.5} & 50.5 & 47.5 & 47.3 & 49.0 \\
ERR (64$\times$64)       & 5.6 & \textbf{51.5} & \textbf{49.0} & \textbf{47.9} & \textbf{50.8} \\

\end{tabular}
}
\vspace{.5em}
  \caption{Ablation study on \emph{enlarging RoI resolution} (ERR) operation, the numbers in brackets are the RoI scales.}
  \label{tab:ablation_err}
\vspace{-.8em}
\end{table}

\begin{table}[t]
\centering
\small
\begin{tabular}{c|cccc}
 & mIoU  &  AP$^\text{p}_\text{50}$ & AP$^\text{p}_\text{vol}$  & PCP$_\text{50}$ \\
 \toprule[0.2em]
baseline           & 50.7 & 47.9 & 47.6 & 49.7 \\
ASPP only        & 51.9 & 51.1 & 48.3 & 51.4 \\
Non-local only  & 50.5 & 47.0 & 47.6 & 48.9 \\
GCE                 & \textbf{52.7} & \textbf{53.2} & \textbf{49.7} & \textbf{52.6} \\

\end{tabular}
\vspace{.5em}
  \caption{Ablation study on \emph{Geometric and Context Encoding} (GCE) module.}
  \label{tab:ablation_gce}
\vspace{-.8em}
\end{table}

\begin{table}[t]
\centering
\small
\scalebox{1.0}{
\begin{tabular}{c|cccc}
 & mIoU  &  AP$^\text{p}_\text{50}$ & AP$^\text{p}_\text{vol}$  & PCP$_\text{50}$ \\
 \toprule[0.2em]
baseline                       & 52.7 & 53.2 & 49.7 & 52.6 \\
4conv + GCE               & 52.8 & 54.9 & 50.5 & 54.2 \\
GCE + 4conv (PBD)    & \textbf{53.5} & 58.5 & \textbf{51.7} & 56.5 \\
4conv + GCE + 4conv & 53.1 & \textbf{58.8} & 51.6 & \textbf{56.7} \\

\end{tabular}
}
\vspace{.5em}
  \caption{Ablation study on \emph{Parsing Branch Decoupling} (PBD) structure. 4conv denotes four convolutional layers with 3$\times$3 kernels.}
  \label{tab:ablation_pbd}
\vspace{-.8em}
\end{table}

\begin{table}[t]
\centering
\small
\begin{tabular}{c|c|cccc}
 & LR & mIoU  &  AP$^\text{p}_\text{50}$ & AP$^\text{p}_\text{vol}$  & PCP$_\text{50}$ \\
 \toprule[0.2em]
\multirow{3}{*}{ImageNet~\cite{Russakovsky_ijcv2015_imagenet}} & 1x & 53.5 & 58.5 & 51.7 & 56.5 \\
                                          & 2x & 55.3 & 61.8 & 53.3 & 59.3 \\
                                          & 3x & \textbf{56.3} & \textbf{63.7} & \textbf{53.9} & \textbf{60.1} \\
\hline
\multirow{3}{*}{COCO~\cite{Lin_eccv2014_coco}}  & 1x & 55.9 & 63.1 & 53.5 & 60.4 \\
                                      & 2x & 57.1 & 64.7 & 54.2 & 61.9 \\
                                      & 3x & \textbf{57.5} & \textbf{65.4} & \textbf{54.6} & \textbf{62.6} \\
\end{tabular}
\vspace{.5em}
  \caption{Results of different pretrained models and maximum iterations on CIHP \texttt{val}.}
  \label{tab:coco_pretrain}
\vspace{-.8em}
\end{table}

\subsection{Implementation Details} 
We implement the Parsing R-CNN based on Detectron on a server with 8 NVIDIA Titan X GPUs. We adopt FPN and RoIAlign in all architectures, each of which is trained end-to-end. A mini-batch involves 2 images per GPU and and each image has 512 sampled RoIs.  We train using image scales randomly sampled from [512, 864] pixels; inference is on a single scale of 800 pixels. For CIHP dataset,  we train on \texttt{train} for 45k iterations, with a learning rate of 0.02 which is decreased by 10 at the 30k and 40k iteration. For MHP v2.0 dataset,  the max iteration is half as long as the CIHP dataset with the learning rate change points scaled proportionally. For DensePose-COCO,  we train for 130k iterations, starting from a learning rate of 0.002 and reducing it by 10 at 100k and 120k iterations. Other details are identical as in Mask R-CNN~\cite{He_iccv2017_maskrcnn, Goyal_arxiv2017_1hour}.

\subsection{Experiments on Human Part Segmentation} 

\begin{table*}[!t]
  \centering
  \small
  \scalebox{0.88}{
  \begin{tabular}{c c c c c c c | c c c c}
     Baseline    & PSS            & ERR           & GCE            & PBD           &  3x LR        & COCO       & mIoU  &  AP$^\text{p}_\text{50}$ & AP$^\text{p}_\text{vol}$ & PCP$_\text{50}$ \\
   \toprule[0.2em]
    ResNet50   &                    &                    &                    &                    &                    &                    &                    &                 \\
    \checkmark &                    &                    &                    &                    &                    &                    &     47.2        &    41.4     &      45.4      &     44.3      \\
    \checkmark & \checkmark &                    &                    &                    &                    &                    &     48.2        &    42.9     &      46.0      &     45.5      \\ 
    \checkmark & \checkmark & \checkmark &                    &                    &                    &                    &     50.7        &    47.9     &      47.6      &     49.7      \\
    \checkmark & \checkmark & \checkmark & \checkmark &                    &                    &                    &     52.7        &    53.2     &      49.7      &     52.6      \\  
    \checkmark & \checkmark & \checkmark & \checkmark & \checkmark &                    &                    &     53.5        &    58.5     &      51.7      &     56.5      \\
    \checkmark & \checkmark & \checkmark & \checkmark & \checkmark & \checkmark &                    &     56.3        &    63.7     &      53.9      &     60.1      \\
    \checkmark & \checkmark & \checkmark & \checkmark & \checkmark & \checkmark & \checkmark &\textbf{57.5} &\textbf{65.4}&\textbf{54.6}&\textbf{62.6} \\
  \bottomrule[0.1em]
  $\Delta$ &     &           &            &           &            &           &   \textbf{+10.3}      &  \textbf{+24.0}  &    \textbf{+9.2}      &    \textbf{+18.3}              \\
  \end{tabular}
 }
  \caption{Human part segmentation results on CIHP \texttt{val}. We adopt ResNet50-FPN as backbone, and gradually add \emph{Proposals separation sampling} (PSS), \emph{Enlarging RoI Resolution} (ERR), \emph{Geometric and Context Encoding} (GCE) and \emph{Parsing Branch Decoupling} (PBD). 3x LR denotes that we increase the number of iterations to three times of standard. We also report the performance of pretraining the whole model on COCO keypoint annotations (COCO). }
  \label{tab:cihp_results}
\end{table*}

\begin{table*}[!t]
  \centering
  \small
  \scalebox{0.88}{
  \begin{tabular}{c c c c c c c | c c c c}
     Baseline    & PSS            & ERR           & GCE            & PBD           &  3x LR        & COCO       & mIoU   &  AP$^\text{p}_\text{50}$& AP$^\text{p}_\text{vol}$  & PCP$_\text{50}$\\
   \toprule[0.2em]
    ResNet50   &                    &                    &                    &                    &                    &                    &                    &                 \\
    \checkmark &                    &                    &                    &                    &                    &                    &     28.7        &    10.1     &      33.4      &     21.8      \\
    \checkmark & \checkmark &                    &                    &                    &                    &                    &     29.8        &    10.6     &      33.8      &     22.2      \\ 
    \checkmark & \checkmark & \checkmark &                    &                    &                    &                    &     32.3        &    14.0     &      34.1      &     27.4      \\
    \checkmark & \checkmark & \checkmark & \checkmark &                    &                    &                    &     33.7        &    17.4     &      36.3      &     30.5     \\
    \checkmark & \checkmark & \checkmark & \checkmark & \checkmark &                    &                    &     34.3        &    20.0     &      37.6      &     32.7      \\
    \checkmark & \checkmark & \checkmark & \checkmark & \checkmark & \checkmark &                    &     36.2        &     24.5    &      39.5      &     37.2      \\
    \checkmark & \checkmark & \checkmark & \checkmark & \checkmark & \checkmark & \checkmark & \textbf{37.0}&\textbf{26.6}&\textbf{40.3}&\textbf{40.0} \\
  \bottomrule[0.1em]
  $\Delta$ &     &           &            &           &            &           &   \textbf{+8.3}       &  \textbf{+16.5}  &    \textbf{+7.1}      &    \textbf{+18.2}               \\
  \end{tabular}
 }
  \caption{Human part segmentation results on MHP v2.0 \texttt{val}, we adopt ResNet50-FPN as backbone.}
  \label{tab:mhp_v2_results}
\end{table*}

\vspace{6pt}
\noindent\textbf{Metrics and Baseline.} We evaluate the performance of human part segmentation from two scenarios. For semantic segmentation, we follow ~\cite{Kirillov_arxiv2018_ps} to generate multi-person mask and adopt the standard mean intersection over union (mIoU)~\cite{Long_cvpr2015_fcn} to evaluate the performance. For instance-level performance, we use the Average Precision based on part (AP$^\text{p}$)~\cite{Zhao_mm2018_mhpv2} for multi-human parsing evaluation, which uses part-level pixel IoU of different semantic part categories within a person instance to determine if one instance is a true positive. We report the AP$^\text{p}_\text{50}$ and AP$^\text{p}_\text{vol}$. The former has a IoU threshold equal to 0.5, and the latter is the mean of the AP$^\text{p}$ at IoU thresholds ranging from 0.1 to 0.9, in increments of 0.1. In addition, we also report Percentage of Correctly parsed semantic Parts (PCP) metric~\cite{Zhao_mm2018_mhpv2}. 

For a fair comparison, our baseline adopts ResNet-50-FPN~\cite{He_cvpr2016_resnet, He_eccv2016_identity, Xie_cvpr2017_resnext, Lin_cvpr2017_fpn} as backbone. The parsing branch consists of a stack of eight 3$\times$3 512-d convolutional layers, followed by a deconvolution~\cite{Zeiler_eccv2014_visualizing} layer and 2$\times$ bilinear upscaling. Following ~\cite{He_iccv2017_maskrcnn}, the feature map resolution after RoIAlign is 14$\times$14, so the output resolution is 56$\times$56.  During training, we apply a per-pixel softmax~\cite{Long_cvpr2015_fcn} as the multinomial cross-entropy loss.

\vspace{6pt}
\noindent\textbf{Component Ablation Studies on CIHP.} We investigate various options of the proposed Parsing R-CNN in Section~\ref{sec:prcnn}. In addition, we also study two other methods to improve performance: increasing the number of iterations and COCO pretraining. Our ablation study on CIHP~\cite{Gong_eccv2018_pgn} \texttt{val} from the baseline gradually to all components incorporated is shown in Table~\ref{tab:cihp_results}. 

\vspace{3pt}
\noindent\textbf{1) Proposals Separation Sampling.} \emph{Proposals separation sampling} (PSS) strategy improves the mIoU about 1.0 than the baseline. We also only adopt the P2 feature map both for bbox branch and parsing branch, the mIoU is reduced by 0.5 and bbox mAP is much worse. As shown in Table~\ref{tab:ablation_pss}, instance-level metrics are promoted to a certain extent with PSS, which indicates that the proposed strategy is effective.

\begin{table}[t]
\centering
\small
\scalebox{0.9}{
\begin{tabular}{c|c|cccc}
 & method & mIoU  &  AP$^\text{p}_\text{50}$ & AP$^\text{p}_\text{vol}$  & PCP$_\text{50}$ \\
 \toprule[0.2em]
\multirow{4}{*}{CIHP}  & PGN (R101)$^\dagger$~\cite{Gong_eccv2018_pgn} & 55.8 & --      & --      & --     \\
                                    & Parsing R-CNN (R50)   & 57.5 & 65.4 & 54.6 & 62.6 \\
                                    & Parsing R-CNN (X101) & 59.8 & 69.1 & 55.9 & 66.2 \\
                                    & Parsing R-CNN (X101)$^\dagger$ & \textbf{61.1} & \textbf{71.2} & \textbf{56.5} & \textbf{67.7} \\
\hline
\multirow{5}{*}{MHP}  & Mask R-CNN~\cite{He_iccv2017_maskrcnn} & -- & 14.9 & 33.8 & 25.1 \\
                                   & MH-Parser~\cite{Li_arxiv2017_mhparser}     & -- & 17.9 & 36.0 & 26.9  \\
\multirow{3}{*}{v2.0}   & NAN~\cite{Zhao_mm2018_mhpv2}               & -- & 25.1 & 41.7 & 32.2  \\ 
                                   & Parsing R-CNN (R50) & 37.0 & 26.6 & 40.3 & 40.0 \\
                                   & Parsing R-CNN (X101) & 40.3 & 30.2 & 41.8 & 44.2 \\
                                   & Parsing R-CNN (X101)$^\dagger$ & \textbf{41.8} & \textbf{32.5} & \textbf{42.7} & \textbf{47.9} \\
\end{tabular}
\vspace{.5em}
}
  \caption{Results of state-of-the-art methods on CIHP and MHP v2.0 \texttt{val}. $^\dagger$ denotes using test-time augmentation.}
  \label{tab:sota_cihp_mhp}
\vspace{-.8em}
\end{table}

\begin{table*}[!t]
  \centering
  \small
  \scalebox{0.88}{
  \begin{tabular}{c c c c c c | c c c | c c}
     Baseline    & PSS            & ERR           & GCE            & PBD           &  COCO        & AP       & AP$_\text{50}$  &  AP$_\text{75}$ & AP$_\text{M}$  & AP$_\text{L}$ \\
   \toprule[0.2em]
    ResNet50   &                    &                    &                    &                    &                    &                    &                    &                 \\
    \checkmark &                    &                    &                    &                    &                    &    48.9         &     84.9        &    50.8     &      43.8      &     50.6      \\
    \checkmark & \checkmark &                    &                    &                    &                    &    50.9         &     86.1        &    53.4     &      46.4      &     52.4      \\ 
    \checkmark & \checkmark & \checkmark &                    &                    &                    &    53.4         &     86.7        &    57.0     &      49.2      &     54.8      \\
    \checkmark & \checkmark & \checkmark & \checkmark &                    &                    &    54.2         &     87.2        &    59.5     &      47.2      &     55.9      \\  
    \checkmark & \checkmark & \checkmark & \checkmark & \checkmark &                    &    55.0         &     87.6        &    59.8     &      50.6      &     56.6      \\
    \checkmark & \checkmark & \checkmark & \checkmark & \checkmark & \checkmark & \textbf{58.3}& \textbf{90.1}& \textbf{66.9}& \textbf{51.8}& \textbf{61.9}\\
    \hline
  $\Delta$ &     &           &            &           &            &   \textbf{+9.4}        &   \textbf{+5.2}       &  \textbf{+16.1}  &    \textbf{+8.0}      &    \textbf{+11.3}               \\
    \bottomrule[0.1em]
   ResNeXt101&                    &                    &                    &                    &                    &                    &                    &                 \\
    \checkmark &                    &                    &                    &                    &                    &     55.5         &     89.1        &    60.8     &      50.7      &     56.8      \\
    \checkmark & \checkmark & \checkmark & \checkmark & \checkmark &                    &    59.1         &     91.0        &    69.4     &      53.9      &     63.1      \\
    \checkmark & \checkmark & \checkmark & \checkmark & \checkmark & \checkmark &\textbf{61.6} &\textbf{91.6} &\textbf{72.3 } &\textbf{54.8}&\textbf{64.8}\\
  \bottomrule[0.1em]
  $\Delta$ &     &           &            &           &            &    \textbf{+6.1}      &   \textbf{+2.5}  &     \textbf{+11.5}      &     \textbf{+4.1}   &  \textbf{+8.0}          \\

  \end{tabular}
 }
  \caption{Dense pose estimation results on DensePose-COCO \texttt{val}. We adopt ResNet50-FPN and ResNeXt101-32x8d-FPN as backbone respectively. The baseline is DensePose-RCNN.} 
  \label{tab:densepose_results}
\end{table*}

\vspace{3pt}
\noindent\textbf{2) Enlarging RoI Resolution.}  In Table~\ref{tab:ablation_err}, we employ 32$\times$32 and 64$\times$64 RoI scales respectively, and find that the performance can be significantly improved than the original 14$\times$14 scale. The ERR (32$\times$32) yields 2.8 improvement in terms of mIoU. For instance-level metrics, the improvements are even greater: 5.0, 1.7, 4.4 respectively. Moreover, the RoIs of paring branch is parallel, so the speed is reduced by only 12\%. And we can increase the inference speed by reducing the number of RoIs. If we use 100 RoIs at inference phase, the speed can be greatly improved and the performance basically does not drop. Relative to 32$\times$32, the 64$\times$64 RoI scale can continue to improve the performance, but considering speed / accuracy trade-offs we consider that using ERR (32$\times$32) is efficient.

\vspace{3pt}
\noindent\textbf{3) Geometric and Context Encoding.} GCE module is the core component of Parsing R-CNN, which can significantly improve the mIoU about 2.0 than stacking of eight 3$\times$3 512-d convolutional layers, and it is even more lightweight. With or without Non-local operation, the ASPP part can still yield 1.2 improvement in terms of mIoU. But without ASPP part, only Non-local operation will cause performance degradation than the baseline. Results are shown In Table~\ref{tab:ablation_gce}.

\vspace{3pt}
\noindent\textbf{4) Parsing Branch Decoupling.} We decouple the parsing branch into three parts: \emph{before GCE},  \emph{GCE module} and  \emph{after GCE}. As shown in Table~\ref{tab:ablation_pbd}, we find that the \emph{before GCE} part is not necessary, and we infer that the GCE module is able to perform semantic space transformation. On the other hand, the \emph{after GCE} part can both significantly improve the semantic segmentation and instance-level metrics (+0.8, +5.3, +2.0, +3.9 respectively). Considering speed / accuracy trade-offs, we adopt the GCE followed by four 3$\times$3 512-d convolutional layers (PBD) as parsing branch.

\vspace{3pt}
\noindent\textbf{5) Increasing iterations and COCO pretraining.} Increasing iterations is a common method for improving performance. As shown in Table~\ref{tab:coco_pretrain}, we investigate the results of twice or three times as long as the standard schedule on CIHP \texttt{val} and find the improvements are obvious. We further pretrain the Parsing R-CNN models on the COCO keypoints annotations\footnote{\fontsize{7pt}{1em}Parsing R-CNN (without ERR) achieves 66.2\% AP on COCO \texttt{val}, which yields 0.8 improvement than Mask R-CNN~\cite{He_iccv2017_maskrcnn} with s1x LR.}, and initialize the parsing branch with the pose estimation weights. This strategy can further improve the performance about 1.1 to 2.4 in terms of mIoU. Combining these two methods, Parsing R-CNN yields 4.0 improvement in terms of mIoU. And for instance-level metrics, the improvements are 6.9, 2.9, 6.1 respectively.

As shown in Table~\ref{tab:cihp_results}, with these proposed components, the metrics of our Parsing R-CNN all exceed the baseline by a big margin. For semantic segmentation, Parsing R-CNN attains 57.6\% mIoU which outperforms the baseline by a massive 10.3 points. For instance-level metrics, the improvement of Parsing R-CNN is more significant, which improves AP$^\text{p}_\text{50}$ by 24.0 points,  AP$^\text{p}_\text{vol}$ by 9.2 points, and PCP$_\text{50}$ by 18.3 points.

\vspace{6pt}
\noindent\textbf{Component Ablation Studies on MHP v2.0.} We also gradually add \emph{Proposals separation sampling} (PSS), \emph{Enlarging RoI Resolution} (ERR), \emph{Geometric and Context Encoding} (GCE) and \emph{Parsing Branch Decoupling} (PBD) for ablation studies on MHP v2.0~\cite{Zhao_mm2018_mhpv2} \texttt{val}, the results are shown in Table~\ref{tab:mhp_v2_results}. There are 59 semantic categories in the MHP v2.0 dataset, and some of them are small-scale, so the baseline is worse than CIHP dataset. Parsing R-CNN is also significantly improving for MHP v2.0 dataset, which yields 10.3 improvement in terms of mIoU. For instance-level metrics, Parsing R-CNN improves AP$^\text{p}_\text{50}$ by 16.5 points,  AP$^\text{p}_\text{vol}$ by 7.1 points, and PCP$_\text{50}$ by 18.2 points.

\vspace{6pt}
\noindent\textbf{Comparisons with State-of-the-Art Methods.} Parsing R-CNN significantly improve the performance of human part segmentation. In order to further prove its effectiveness, we compare the proposed Parsing R-CNN to the state-of-the-art methods on CIHP and MHP v2.0 datasets, respectively.

For CIHP dataset, Parsing R-CNN using ResNet-50-FPN outperforms the PGN~\cite{Gong_eccv2018_pgn} which using ResNet-101 by 1.7 points in terms of mIoU (Table~\ref{tab:sota_cihp_mhp}). It is worth noting that PGN adopts multi-scale inputs and left-right flipped images to improve performance, while the result of Parsing R-CNN is without test-time augmentation. We also report the performance of Parsing R-CNN using ResNeXt-101-32x8d-FPN backbone, which attains 59.8\% mIoU. Moreover, using ResNeXt-101-32x8d-FPN we report the results with multi-scale testing and horizontal flipping. This gives us a single model result of 61.1\% mIoU. Because PGN only reports the Average Precision based on region (AP$^\text{r}$), we can not directly compare the instance-level metrics. But by the result of semantic segmentation, we can also infer that Parsing R-CNN is superior to PGN on human parts segmentation task.

For MHP v2.0 dataset, we also report the results of Parsing R-CNN using ResNet-50-FPN and ResNeXt-101-32x8d-FPN (with or without test-time augmentation) backbones. In Table~\ref{tab:sota_cihp_mhp}, compared with the previous state-of-the-art methods~\cite{He_iccv2017_maskrcnn, Li_arxiv2017_mhparser, Zhao_mm2018_mhpv2}\footnote{\fontsize{7pt}{1em}All the previous state-of-the-art methods only report the results evaluated on MHP v2.0 \texttt{test}.}, Parsing R-CNN further improves results, with a margin of 7.4 points AP$^\text{p}_\text{50}$, 1.0 points AP$^\text{p}_\text{vol}$ and 15.7 points PCP$_\text{50}$ over the best previous entry. Unfortunately, all the methods do not give the metric of semantic segmentation.


\begin{table}[t]
\centering
\small
\scalebox{0.95}{
\begin{tabular}{c|ccc|cc}
 & AP       & AP$_\text{50}$  &  AP$_\text{75}$ & AP$_\text{M}$  & AP$_\text{L}$ \\
 \toprule[0.2em]
 DensePose-RCNN        & 56 & 89 & 64 & 51 & 59 \\
 yuchen.ma                     & 57 & 87 & 66 & 48 & 61 \\
 ML-LAB                          & 57 & 89 & 64 & 51 & 59 \\
 Min-Byeonguk               & 58  & 89 & 66 & 50 & 61 \\
 \textbf{Parsing R-CNN (ours)} & \textbf{64}  & \textbf{92} & \textbf{75} & \textbf{57} & \textbf{67} \\
\end{tabular}
}
\vspace{.5em}
  \caption{2018 COCO Challenge results of Dense Pose Estimation task on \texttt{test}.}
  \label{tab:coco2018_densepose}
\vspace{-.8em}
\end{table}

\begin{figure*}
\begin{center}
\includegraphics[width=0.9\linewidth]{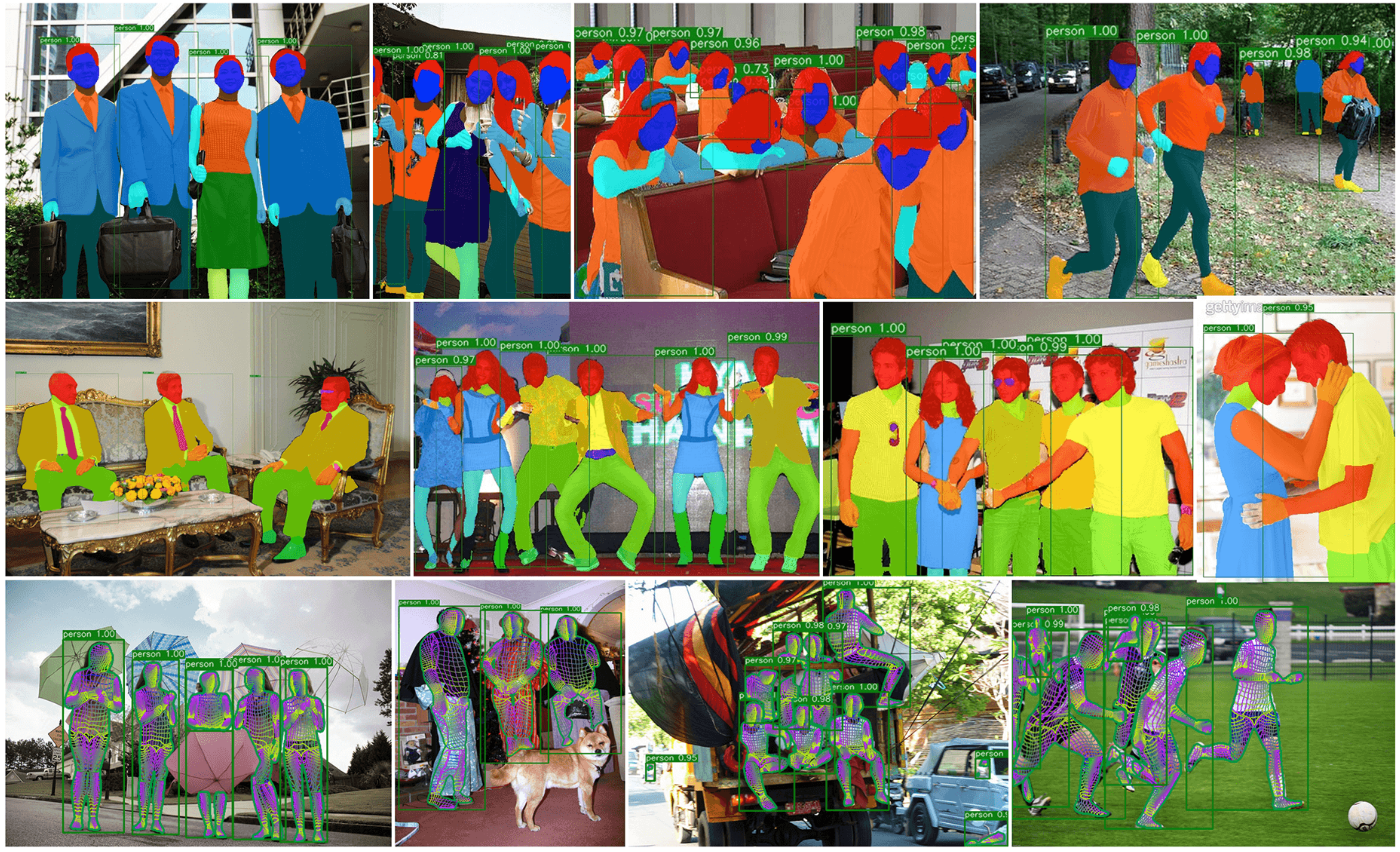}
\end{center}
\vspace{-1mm}
\caption{Images in each row are visual results of Parsing R-CNN using ResNet50-FPN on CIHP \texttt{val}, MHP v2.0 \texttt{val} and DensePose-COCO \texttt{val}, respectively.}
\label{fig:vis_results}
\end{figure*}

\subsection{Experiments on Dense Pose Estimation} 
\vspace{6pt}
\noindent\textbf{Metrics and Baseline.} Following ~\cite{Guler_cvpr2018_densepose}, we adopt the Average Precision (AP) at a number of \emph{geodesic point similarity} (GPS) thresholds ranging from 0.5 to 0.95 as the evaluation metric. The structure of baseline model is exactly the same as the one of human part segmentation. We only replace the per-pixel softmax loss with the dense pose estimation losses.

\vspace{6pt}
\noindent\textbf{Component Ablation Studies on DensePose-COCO.} Like human part segmentation, we adopt the proposed Parsing R-CNN for dense pose estimation. Corresponding results are shown in Table~\ref{tab:densepose_results}. We adopt ResNet50-FPN and ResNeXt101-32x8d-FPN as backbone respectively. With ResNet50-FPN, Parsing R-CNN outperforms the baseline (DensePose-RCNN) by a good margin. Combining all the proposed components, our method achieves 55.0\% AP, which yields 6.1 improvement than DensePose-RCNN. With COCO pretraining, Parsing R-CNN further improves 3.3 points AP. Parsing R-CNN also shows significant improvement of AP$_\text{75}$ (50.8\% vs 66.9\%), which indicates that our method is more accurate in points localization on the surface. As shown in Table~\ref{tab:densepose_results}, our Parsing R-CNN still increase the performance of dense pose estimation, when  the model is upgrade from ResNet50 to ResNeXt101-32x8d, showing good generalization of the Parsing R-CNN framework.

\vspace{6pt}
\noindent\textbf{COCO 2018 Challenge.} With Parsing R-CNN, we participated in the COCO 2018 DensePose Estimation Challenge, and reach the 1st place over all competitors. Table~\ref{tab:coco2018_densepose} summarizes the entries from the leaderboard of COCO 2018 Challenge. Our entry only utilizes a single model (ResNeXt101-32x8d), and attains 64.1\% AP on DensePose-COCO \texttt{test} which surpasses the 2nd place by 6 points.

Qualitative results are illustrated in Figure~\ref{fig:vis_results}. Images in each row are visual results of Parsing R-CNN using ResNet50-FPN on CIHP \texttt{val}, MHP v2.0 \texttt{val} and DensePose-COCO \texttt{val}, respectively. 


\section{Conclusion} We present a novel region-based approach Parsing R-CNN for instance-level human analysis, which achieves state-of-the-art results on several challenging benchmarks. Our approach explores the problem of instance-level human analysis from four aspects, and verified the effectiveness on human part segmentation and dense pose estimation tasks. Based on the proposed Parsing R-CNN, we reach the 1st place in the COCO 2018 Challenge DensePose Estimation task. In the future, we will extend Parsing R-CNN to more applications of instance-level human analysis.

{\small
\bibliographystyle{ieee}
\bibliography{egbib}
}

\end{document}